\documentclass[conference]{IEEEtran}
\IEEEoverridecommandlockouts
\usepackage{cite}
\usepackage{amsmath,amssymb,amsfonts}
\usepackage{algorithmic}
\usepackage{algorithm}
\usepackage{graphicx}
\usepackage{textcomp}
\usepackage{xcolor}
\usepackage{arydshln}
\usepackage{subfig}
\usepackage{breqn}
\usepackage{csquotes}
\usepackage{booktabs}
\usepackage{multirow}
\def\BibTeX{{\rm B\kern-.05em{\sc i\kern-.025em b}\kern-.08em
    T\kern-.1667em\lower.7ex\hbox{E}\kern-.125emX}}
\begin{document}
\newcommand{\armar}{\IEEEauthorrefmark}
\newcommand{\etal}{\textit{et al.}}

\title{Iterative, Deep, and Unsupervised \\ Synthetic Aperture Sonar Image Segmentation
}

\author{
    \IEEEauthorblockN{Yung-Chen Sun\armar{1}, Isaac D. Gerg\armar{1}\armar{2}, Vishal Monga\armar{1}}
    \IEEEauthorblockA{\armar{1}Department of Electrical Engineering, Pennsylvania State University, University Park, PA \\
    \armar{2}Applied Research Laboratory, Pennsylvania State University, State College, PA}
\vspace{-0.5cm}
}

\maketitle

\begin{abstract}
Deep learning has not been routinely employed for semantic segmentation of seabed environment for synthetic aperture sonar (SAS) imagery due to the implicit need of abundant training data such methods necessitate. Abundant training data, specifically pixel-level labels for all images, is usually not available for SAS imagery due to the complex logistics (e.g., diver survey, chase boat, precision position information) needed for obtaining accurate ground-truth. Many hand-crafted feature based algorithms have been proposed to segment SAS in an unsupervised fashion. However, there is still room for improvement as the feature extraction step of these methods is fixed. In this work, we present a new iterative unsupervised algorithm for learning deep features for SAS image segmentation. Our proposed algorithm alternates between clustering superpixels and updating the parameters of a convolutional neural network (CNN) so that the feature extraction for image segmentation can be optimized. We demonstrate the efficacy of our method on a realistic benchmark dataset.  Our results show that the performance of our proposed method is considerably better than current state-of-the-art methods in SAS image segmentation.

% Synthetic aperture sonar (SAS) systems produce high-resolution images of the seabed environment. Moreover, deep learning has demonstrated superior ability in finding robust features for automating imagery analysis. However, the success of deep learning is conditioned on having lots of labeled training data, but obtaining generous pixel-level annotations of SAS imagery is often practically infeasible. This challenge has thus far limited the adoption of deep learning methods for SAS segmentation. Algorithms exist to segment SAS imagery in an unsupervised manner, but they lack the benefit of state-of-the-art learning methods and the results present significant room for improvement. In view of the above, we propose a new iterative algorithm for unsupervised SAS image segmentation combining superpixel formation, deep learning, and traditional clustering methods. A comparison to current state-of-the-art methods on a realistic benchmark dataset for SAS image segmentation demonstrates the benefits of our proposal. 

% Because our design combines merits of classical superpixel methods with deep learning, practically we demonstrate a very significant benefit in terms of reduced selection bias, i.e. IDUS shows markedly improved robustness against the choice of training images. 
% Finally, we also develop a semi-supervised (SS) extension of IDUS called IDSS and demonstrate experimentally that it can further enhance performance while outperforming U-Net based supervised deep learning methods that exploit the same labeled training imagery. 
\end{abstract}


\section{Introduction}
% paper contribution, total section should be one page with abstract

Recently, deep learning frameworks applied for image segmentation have shown promising results \cite{7440871,unet}. To segment distinct seabed textures in sonar images, a naive intuition is that we can use a similar network structure for seabed segmentation of SAS images because they have largely composted low-level textures. However, training a convolutional neural network (CNN) requires a considerable amount of labeled data to obtain good results. For SAS image segmentation, this means acquiring pixel-level annotations which is a costly process, and rarely do we have complete ground-truth from a diver survey. Consequently, humans often label regions which are easily recognized while forgoing areas difficult to distinguish such as region boundaries or blended regions. As a result, few pixel-level labeled datasets are available for training.
\begin{figure}[t]
    \centering
    \begin{tabular}{cc}
        \includegraphics[height=4cm]{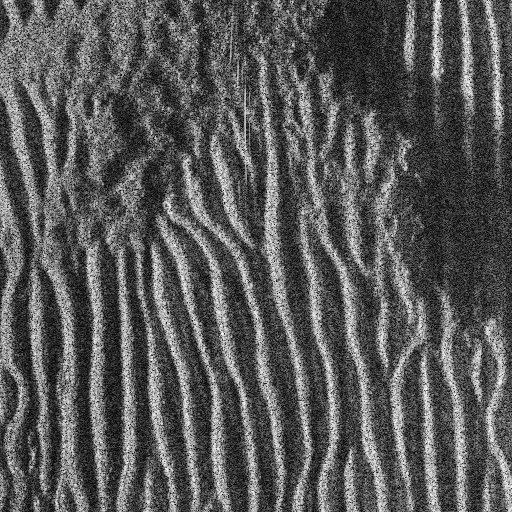}&\includegraphics[height=4cm]{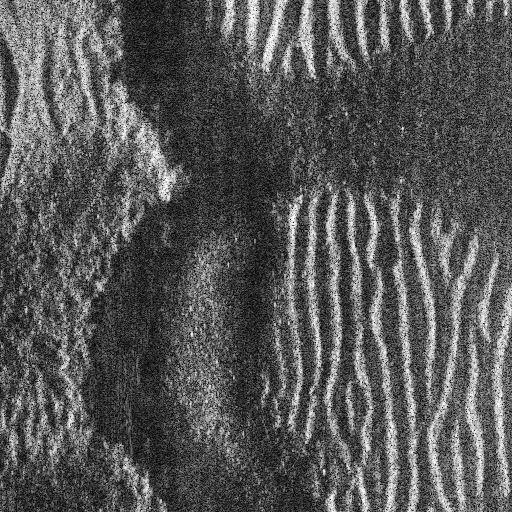}
    \end{tabular}
    \caption{Sample SAS images containing various seabed types including large sand ripples and flat sandy seafloor.}
    \label{fig:sonar_example}
 \end{figure}
 
Unsupervised image classification \cite{caron2018deep} has shown that embedding deep learning steps into a clustering algorithm is a promising direction. The essence of image-level clustering is to see every image as a region that only contains one high-level semantic texture, but the regions’ boundaries are inaccurate. We apply this idea to the image segmentation task. Although we are going to do pixel-level segmentation, from our observations regions containing a single texture will produce similar pixel predictions. If we can find accurate boundaries to differentiate the regions containing a single texture, then the image segmentation process can be transferred into the region-level classification problem.  

We devise a novel unsupervised segmentation method that combines deep learning and an iterative clustering technique producing an algorithm suitable for unsupervised seabed segmentation of SAS images. The basic idea is to use superpixels generated by CNN pseudo-labels to split one sonar image into many small regions. Each region is quantized into one superpixel feature that is the average softmax value and then a clustering process is performed to group these regions. Finally, we utilize the clustering results to update the CNN parameters to encourage more accurate predictions of a segmentation network.

We evaluate our method, which we call iterative deep unsupervised segmentation (IDUS), against several comparison methods on a contemporary real-world SAS dataset.  We show compelling performance of IDUS against existing state-of-the-art methods for SAS image co-segmentation.  Additionally, we show an extension of our method which can obtain even better results when semi-labeled data is present.
%Besides this, the benefit of combining IDUS with supervised training is also be demonstrated to explore possible applications for the real-world case. 
\textbf{Specifically, our work makes two main contributions:}
\begin{enumerate}
    \item \textbf{We devise a new iterative method to integrate the merits of best-known unsupervised superpixel segmentation along with a deep feature extractor.} 
    % In particular, we utilize superpixels to generalize image-level clustering to superpixel clustering and propose an unsupervised learning framework for training a U-Net based CNN \cite{unet} in image segmentation tasks. Our proposal exceeds state-of-the-art performance on the task of unsupervised segmentation for seabed sonar images.
    % \item We propose a supervised extension for IDUS, which called \textbf{Iterative Deep Semi-supervised Segmentation (IDSS)}, to increase network performance in situations that do not have enough annotated data.
    % \item IDUS is \textbf{considerably faster} than the previously widely used methods, which encourages the deployment of IDUS on applications that require real-time speed.
    \item \textbf{Our approach does not depend on any specific network structure or superpixel generation algorithm.} The user can adjust the algorithm easily according to different segmentation tasks.
\end{enumerate}

\section{Technical Approach}
% whole section should be one page
\begin{figure*}[t]
\setlength{\belowcaptionskip}{-0.5cm} 
    \centering
    \begin{tabular}{c}        
        \includegraphics[width=0.9\linewidth]{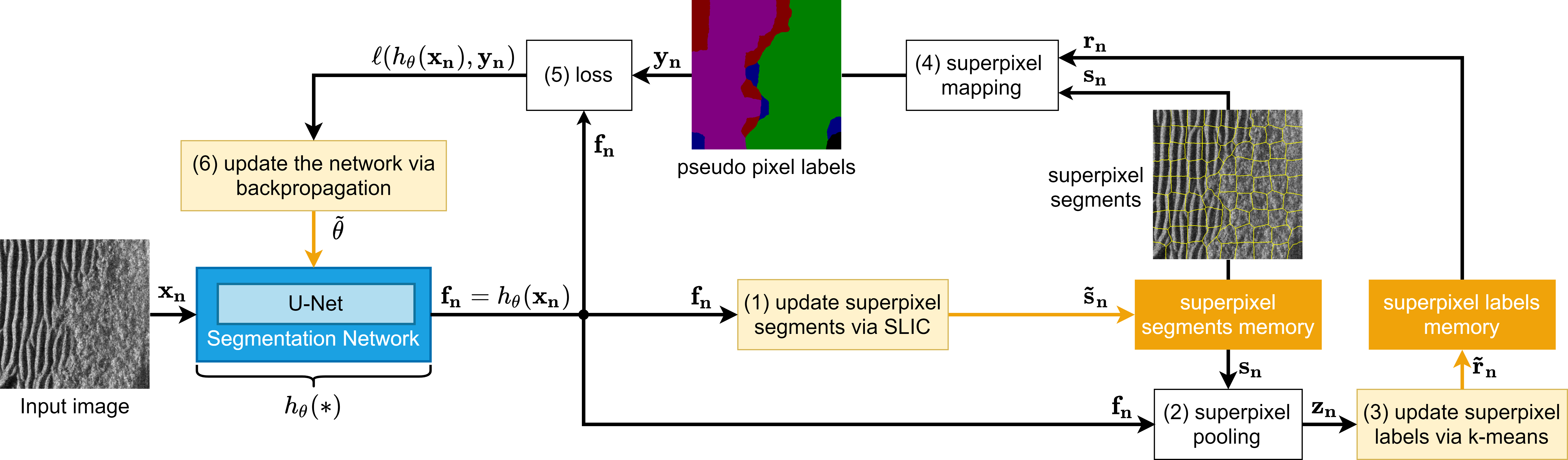}
    \end{tabular}
    \caption{
        \small
        An iteration of IDUS: (1) Use SLIC \cite{SLIC} to generate the new superpixel segments. (2) Quantize pixel features into superpixel features. (3) Cluster superpixel features by k-means to update the class assignments in memory. (4) Map superpixel labels to pseudo pixel labels. (5) Compute the loss via the pseudo pixel labels and the network softmax outputs. (6) Update the network via backpropagation.
    }
    \label{fig:IDUS}
\end{figure*}
A recent method called DeepCluster \cite{caron2018deep} demonstrates that the combination of clustering techniques and CNNs show promise in unsupervised learning of deep features on image classification dataset is possible. The benefit of DeepCluster is that the training method is more generally and easily transferred to other tasks. This benefit motivates us to develop our own work on unsupervised image segmentation because the method not only will not be limited to the sonar images, but it can also be used in other segmentation tasks.

\textbf{Notation.} Given a set $\mathbf{X}=\left\{\mathbf{x_1},...,\mathbf{x_N}\right\}$ of $N$ SAS images, $s_n^k$ is the $k$-th superpixel of the $n$-th image and $r_n^k$ are the pseudo-label of this superpixel. We use $h_{\theta}(\cdot)$ to denote the softmax output of a segmentation network (i.e. deep network), where $\theta$ is the parameters we need to optimize. The output  has same size with the input image. Given an input image $\mathbf{x_n}$, the network maps the image into a feature map $\mathbf{f_n}=h_{\theta}(\mathbf{x_n})$ in which $f_n^i$ represent the feature of $i$-th pixel $x_n^i$ in this image.

\textbf{Superpixel Pooling.} To quantize the pixel-level features into region-level features $\mathbf{z_n}=Q_{pool}(\mathbf{x_n},\mathbf{f_n},\mathbf{s_n})$, we compute the feature's mean from all the pixels making up the  superpixel $z_n^k$ by the following equation: 
\begin{equation}\label{eq:1}
   z_n^k=\frac{1}{D}\sum_{i=1}^{D}f_n^iI(x_n^i \in s_n^k),
\end{equation}
where $D$ denotes the number of pixels in the input image. $I(x_n^i \in s_k^i)$ is an indicator function that is 1 if $x_n^i$ belongs to the superpixel $s_n^k$. The superpixel quantization makes sure that we can collect regional features of sufficient accuracy for the later clustering use.

\textbf{Superpixel Mapping.} Suppose we have already known a superpixel label $r_n^k$, then every pixel label $y_n^i$ in this superpixel should be the same. We use the following formula to describe this mapping:
\begin{equation}\label{eq:2}
    y_n^i=\sum_{k=1}^{K_n}r_{n}^{k}I(x_n^i \in s_n^k)
\end{equation}
where $K_n$ denotes the number of superpixels in $n$th image. We use $\mathbf{y_n} = Q_{map}(\mathbf{x_n},\mathbf{r_n},\mathbf{s_n})$ to denote this process which we call superpixel mapping. It eliminates some classification errors and produces more robust class assignments for pixels.

\textbf{The Simplest Version of IDUS.} \figurename~\ref{fig:IDUS} shows an iteration of the simplest version of IDUS after initialization. Here, we assume that the minibatch only contains one image. The superpixel segments memory stores the boundary of every superpixel and the superpixel labels memory stores the superpixel labels. We see the softmax vectors of network outputs as a kind of feature representation, due to it has the same size of the inputs and contains the distance information between a region texture and cluster centroids. In an iteration, we first use superpixel segments to quantize the pixel features into superpixel features (Eq. \eqref{eq:1}) and use $k$-means to update the superpixel labels. Next, we map the superpixel labels to pixel labels (Eq. \eqref{eq:2}) and use the pseudo-class assignments to update the network via back-propagation. As shown in \figurename~\ref{fig:IDUS}, IDUS alternates between generating pseudo ground-truth labels and updating the network parameters.

\textbf{The Complete Version of IDUS} Given a set of training images $\mathbf{X_T} = \left\{\mathbf{x_n}\right\}_{n=1}^{N_{T}}$, we update the model's parameters $\theta$ by solving the following problem,
\begin{equation}\label{eq:3}
    \min_{\theta}\frac{1}{N_{T}}\sum_{n=1}^{N_{T}}\ell(h_{\theta}(\mathbf{x_n}), Q_{map}(\mathbf{x_n},\mathbf{r_n},\mathbf{s_n}) ),
\end{equation}
where $\ell$ is a loss function and $\mathbf{r_n}$, $\mathbf{s_n}$ will be initialized before performing the algorithm. After training the model for $U_{E}$ epochs, we extract the feature maps $\left\{\mathbf{f_n}\right\}_{n=1}^{N_T}$ of all training images based on the current model's parameters $\theta_{i}$ where $i$ is the number of iteration. Then, we pool the feature maps to regional features $\left\{\mathbf{z_n}\right\}_{n=1}^{N_T}$ and cluster them into $M$ groups based on the optimization goal
\begin{equation}\label{eq:4}
    \min_{C\in\mathbb{R}^{M\times d}}\sum_{m=1}^M\sum_{n=1}^N\sum_{k=1}^{K_n}\min_{r_n^k \in [1, M]}\|z_n^k-c_m\|I(r_n^k==m),
\end{equation}
where $c_m$ is the centroid feature of class $m$ and $d$ is the length of feature dimension. With this clustering criterion, the regional features will be clustered and used to produced new labels $\left\{\mathbf{r_n}\right\}_{n=1}^{N_T}$; and then, they will be assigned to the corresponding superpixels.

The overall idea of IDUS can be summarized in Algorithm \ref{alg:algorithm1}. We update the superpixel boundaries every $U_{S}$ epochs and the algorithm terminates after a fixed number of iterations. 

\begin{algorithm}
    \caption{Iterative Deep Unsupervised Segmentation}
    \begin{algorithmic}[1]
        \STATE Initialize superpixel segments, labels $\left\{\mathbf{s_n},\mathbf{r_n}\right\}_{n=1}^{N_T}$ of $N_T$ training images. And map the regional labels to pseudo pixel labels $\left\{\mathbf{y_n}\right\}_{n=1}^{N_T}$ by Eq \ref{eq:2}.
        \STATE Assign intervals $U_{E}$, $U_{S}$ for updating the pseudo-ground-truth and superpixel boundaries.
    \REPEAT
        
        \STATE Given training data set $\left\{\mathbf{x_n}, \mathbf{y_n}\right\}_{n=1}^{N_{T}}$, update the model's parameters $\theta$ based on the problem Eq \eqref{eq:3}.
        \IF{every $U_S$ epochs}
            \STATE Update superpixel boundaries $\left\{\mathbf{s_n}\right\}_{n=1}^N$ based on the model of current iteration.
        \ENDIF
        \IF{every $U_E$ epochs}
            \STATE Use the model parameters of current iteration $\theta_{i}$ to generate pixel-level features $\{\mathbf{f_n}\}_{n=1}^{N_T}$.
            \STATE Pool the pixel-level features into regional features by Eq \eqref{eq:1}.
            \STATE Cluster the regional features based on criterion Eq \eqref{eq:4} and reassign new pseudo labels $\left\{\mathbf{r_n}\right\}_{n=1}^{N_T}$ to each superpixel.
        \ENDIF
    \UNTIL{reach the maximum iteration.}
    \STATE Output the model's parameters $\theta$.
    \end{algorithmic}
    \label{alg:algorithm1}
\end{algorithm}

\begin{table}
    \centering
    \begin{tabular}{ccccc}
    \toprule
    Layer Name & Layer Function & Dim. & \# Filters & Input \\\midrule
    up1        & Upsampling     & 2$\times$2   & N/A      & res4 \\
    merge1     & Concatenate    & N/A          & N/A      & up1, res3 \\
    conv1a     & Conv+BN+ReLU    & 3$\times$3   & 256      & merge1 \\
    conv1b     & Conv+BN+ReLU     & 3$\times$3   & 256      & conv1a \\\midrule
    
    up2        & Upsampling     & 2$\times$2   & N/A      & conv1b\\
    merge2     & Concatenate    & N/A          & N/A      & up2, res2 \\
    conv2a     & Conv+BN+ReLU     & 3$\times$3   & 128      & merge2 \\
    conv2b     & Conv+BN+ReLU     & 3$\times$3   & 128      & conv2a \\\midrule
    
    up3        & Upsampling     & 2$\times$2   & N/A      & conv2b\\
    merge3     & Concatenate    & N/A          & N/A      & up3, res1 \\
    conv3a     & Conv+BN+ReLU     & 3$\times$3   & 64       & merge3 \\
    conv3b     & Conv+BN+ReLU     & 3$\times$3   & 64       & conv3a \\\midrule
    
    up4        & Upsampling     & 2$\times$2   & N/A      & conv3b\\
    merge4     & Concatenate    & N/A          & N/A      & up4, res\_conv \\
    conv4a     & Conv+BN+ReLU     & 3$\times$3   & 32       & merge4 \\
    conv4b     & Conv+BN+ReLU     & 3$\times$3   & 32       & conv4a \\\midrule
    
    up5        & Upsampling     & 2$\times$2   & N/A      & conv4b\\
    conv5a     & Conv+BN+ReLU     & 3$\times$3   & 16       & up5 \\
    conv5b     & Conv+BN+ReLU     & 3$\times$3   & 16       & conv5a \\\midrule
    
    seg\_head   & Conv+Softmax     & 3$\times$3   & 7        & conv5b \\
    \bottomrule
    \end{tabular}
\caption{Description of decoder and segmentation head of the U-Net used in IDUS.  This decoder is preceded by a pre-trained ResNet-18 network.  ``Conv+BN+ReLU" is 2D convolutional followed by Batch Normalization followed by rectified linear unit (ReLU) activation and ``Softmax" denotes the softmax function. We use layers res\_conv, res1, res2, res3, and res4 to denote the output of ResNet-18's first convolutional layer and building blocks, respectively. Layer conv5b is the decoder output and layer seg\_head is the segmentation head output.} 
\label{tab:network_struct}
\end{table}

\textbf{Implementation Details}
The backbone model used by IDUS is U-Net\cite{unet} equipped with ResNet-18 encoder that pre-trained on ImageNet \cite{imagenet_cvpr09}. \tablename~\ref{tab:network_struct} shows the decoder and segmentation head (pixel-level linear classifier) of U-Net. For the initialization, we first concatenate the feature maps of the output of the first and third convolutional blocks in ResNet-18 of every image; and then, we use the idea of texton \cite{malik2001contour} to initialize the pseudo labels and a two-step texton selection \cite{Cobb2013MultiimageTS} to reduce the computation time. We train the IDUS with mini-batch size fifteen on an NVIDIA Titan X GPU (12GB) with the PyTorch package \cite{NEURIPS2019_bdbca288}. The loss function we use is the mean of dice loss \cite{sudre2017generalised} and cross-entropy loss. We use weights $w_m^{ept} = 1/r_m$ and $w_m^{dice} = 1/\sqrt{r_m}$, where $r_m$ was the proportion of $m$th class label in training samples, to balance the class weights of two losses, respectively. Batch normalization layers \cite{ioffe2015batch} are added after each convolutional layer (except the segmentation head) to accelerate the loss convergence. The model's parameters are initialed by a uniform distribution following the scaling of \cite{he2015delving} and optimized by Adam \cite{kingma2014adam} algorithm with weight decay value $10^{-9}$. In every iteration, the learning rate of the model is initialized at $10^{-4}$ and decreased by $0.1$ every fifty epochs. The updating interval of superpixel labels and boundaries is 200 and we do this five times throughout training. Consequently, we train for 1000 epochs where every 200 epochs we update the superpixel labels and boundaries.

\section{Experimental Results}
\subsection{Experimental Setup}

The dataset we use to test the performance of IDUS is composed of 113 images collected from a high-frequency SAS system which are semi-labeled (i.e. not every pixel is labeled in every image). The images contain seven distinct labeled seabed areas: shadow (SH), dark sand (DS), bright sand (BS), seagrass (SG), large sand ripple (SR), rock (RK), and small sand ripple (SR). The images were originally $1001 \times 1001$ pixels in size which we downsample to $512\times512$ pixels for computational efficiency.

\textbf{Comparison Methods.} To show the effectiveness of our method, we compare with two recent methods \cite{lianantonakis2007sidescan, Zare2017PossibilisticFL} used in SAS seabed environment image segmentation. Following, we give the implementation details we use in generating the comparison methods as no source code is publicly available to evaluate. We make a best effort attempt to reproduce the methods as given in their respective sources. 

\textbf{Lianantonakis, \textit{et al.} (2007) \cite{lianantonakis2007sidescan}.} This method uses Haralick features \cite{haralick1973textural} derived from the gray-level co-occurrence matrix (GLCM) and couples this with active contours to arrive at a binary class mapping. 
We extend this work to multiple classes by simply using the same feature descriptors as the original work but apply $k$-means++ to cluster; a similar replication approach is used in \cite{cobb2011autocorrelation}. We ran $k$-means++ with 100 random initializations and selected the run that produces the minimum within cluster sum of squares error in a manner consistent with \cite{williams2009unsupervised}. 

\textbf{Zare, \textit{et al.} (2017) \cite{Zare2017PossibilisticFL}.} In this work, the feature sets are produced by Sobel edge descriptors (Sobel) \cite{4610973}, histograms of oriented gradients (HOG) \cite{dalal2005histograms}, and local binary pattern (LBP) features \cite{guo2010completed}. For each feature descriptor, we use the same sliding window strategy of Lianantonakis, \etal \cite{lianantonakis2007sidescan} to derive a feature vector for each pixel.

For the comparison methods, the co-segmentation strategy contains four main steps:
\begin{enumerate}
\item Feature extraction to extract a set of feature maps from the input sonar images. 
\item Performing SLIC \cite{SLIC} on the extracted features to generate superpixel features by Eq \eqref{eq:1}. 
\item Unsupervise cluster all the superpixels and provide each superpixel a cluster assignment. 
\item Pixels inside a superpixel are assigned the same cluster by Eq \eqref{eq:2} to get pixel-level co-segmentation.
\end{enumerate}

For the co-segmentation strategy of IDUS, we train a segmentation network by IDUS on the entire dataset first, and then directly use the network outputs computed from the input sonar images as the final co-segmentation results.

\subsection{Results}
\begin{figure}
    \centering
    \setlength{\belowcaptionskip}{-0.5cm}
    \begin{tabular}{c}
        \includegraphics[width=1.1\linewidth]{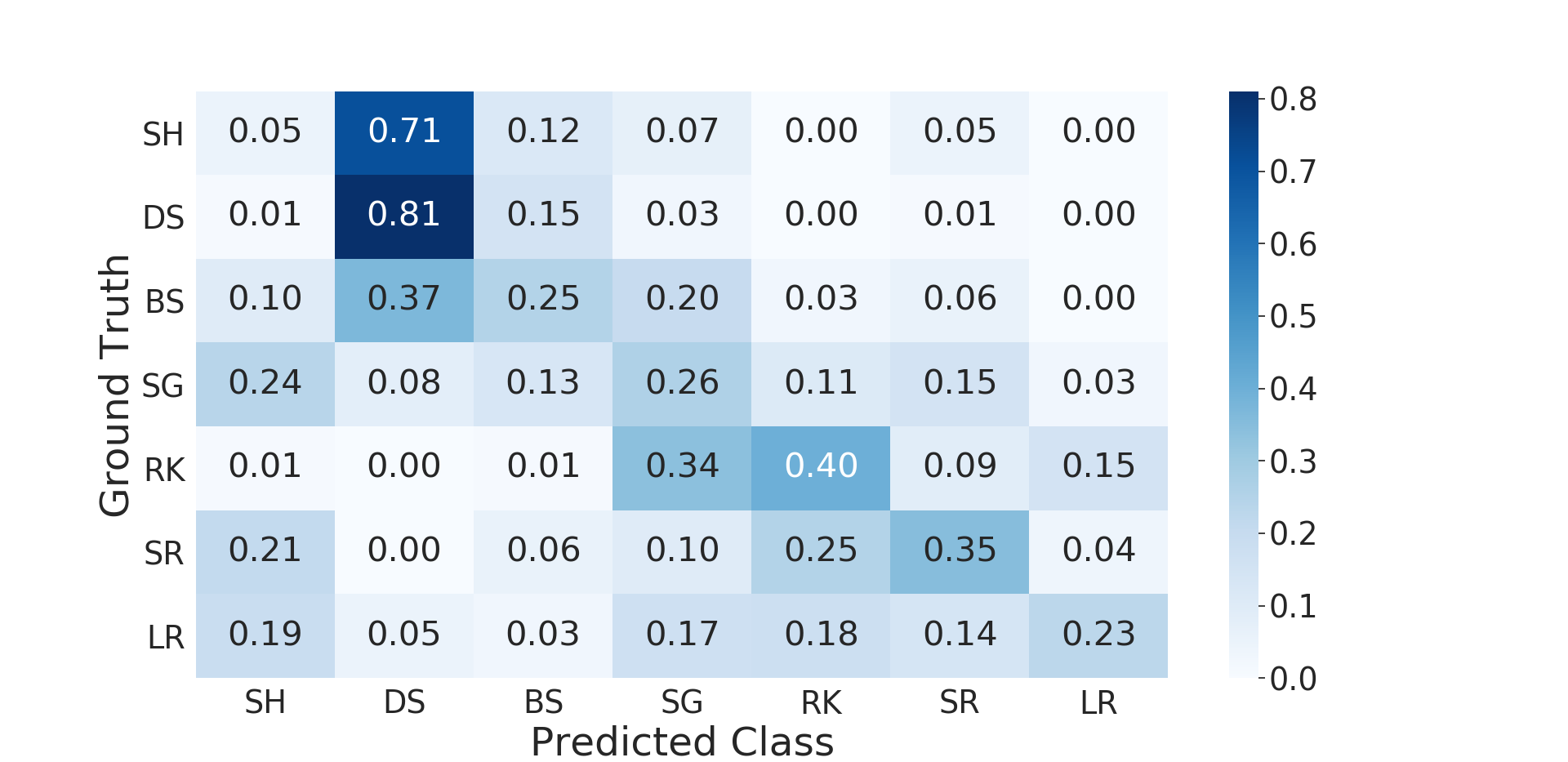} \\
        (a) Lianantonakis, \textit{et al.} (2007) \cite{williams2009unsupervised} 2009 \\
        \includegraphics[width=1.1\linewidth]{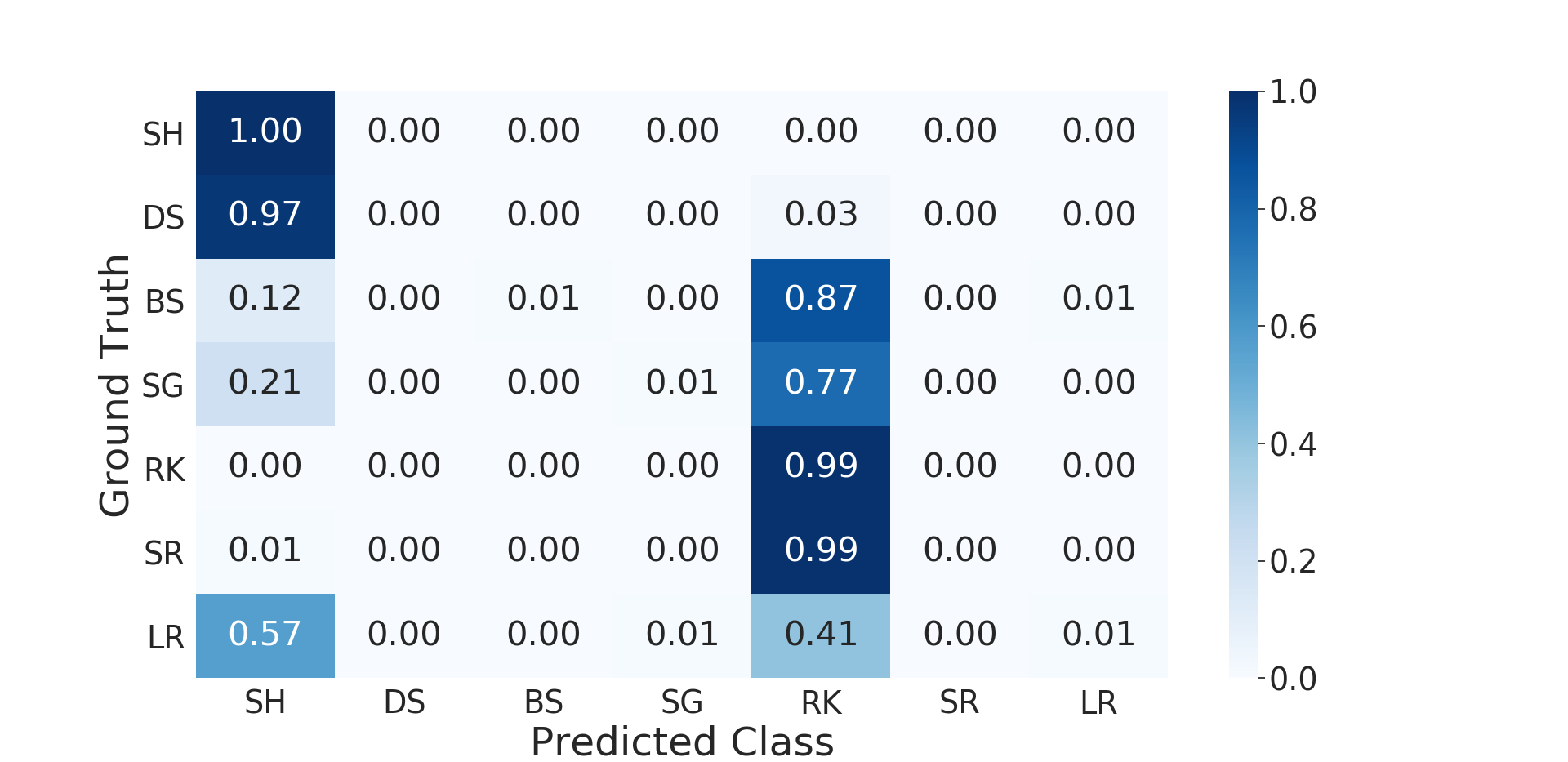} \\
        (b) Zare, \textit{et al.} (2017) \cite{Zare2017PossibilisticFL} 2017\\
        \includegraphics[width=1.1\linewidth]{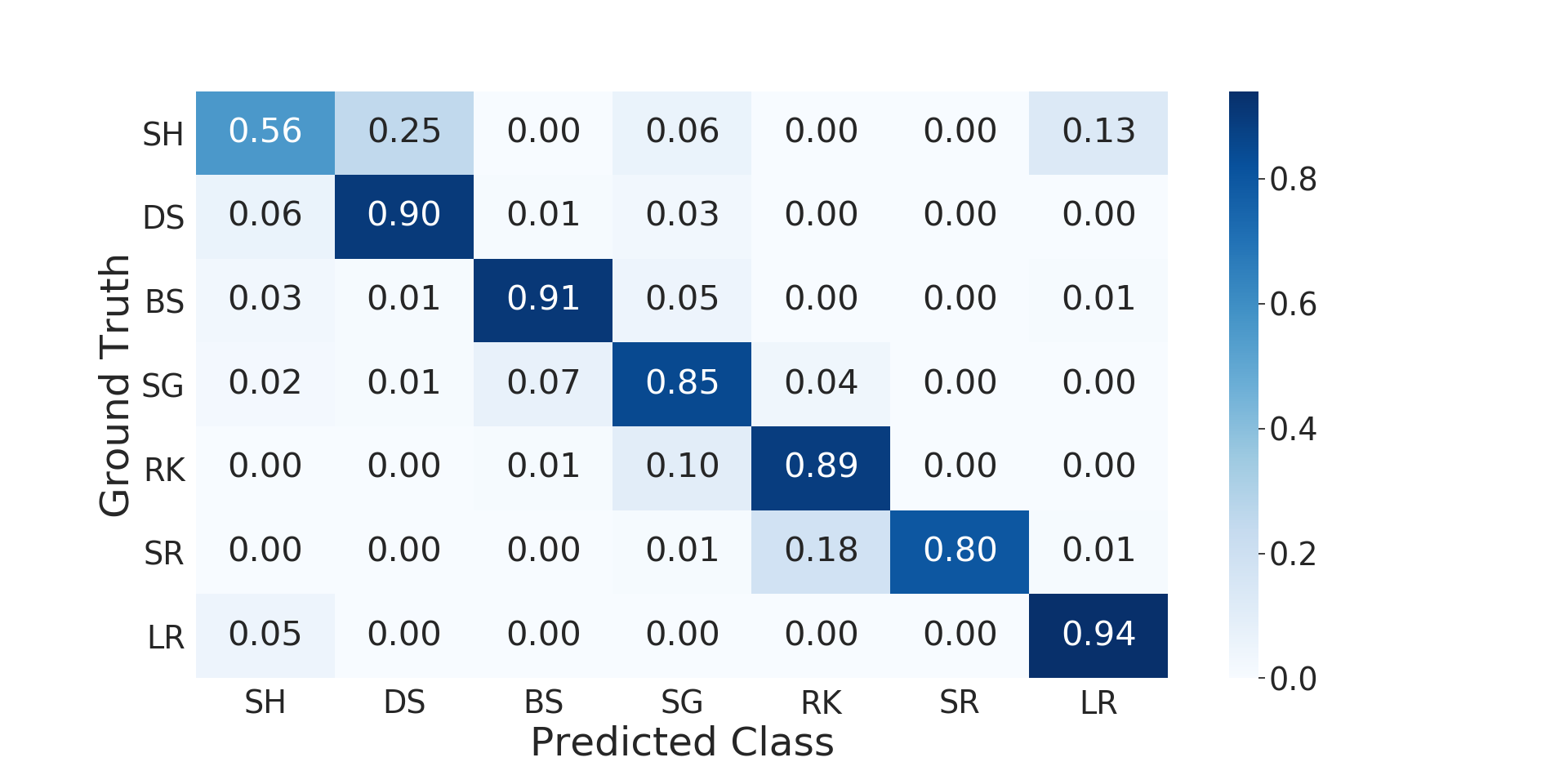} \\
        (c) IDUS
    \end{tabular}
    \caption{Confusion matrices of IDUS along with comparisons in the first experiment. Higher values along the diagonal indicate better performance.}
    \label{fig:confusion_matrix}  
\end{figure}
We use the confusion matrix as a criterion to evaluate the co-segmentation performance of IDUS and the comparison methods on the entire dataset (113 images). A confusion matrix shows the proportion of an assigned class predicted by the algorithm for a given ground-truth class. Ideally, the ground-truth class and the predicted class overlap entirely giving a proportion of one. However, in practice, the predicted class by co-segmentation usually do not perfectly match with the ground-truth class resulting in a proportion less than one. 

As a result of performing unsupervised segmentation, class assignments by the algorithm must be mapped to ground-truth classes. We perform this assignment by initially using a random disjoint assignment among the possible classes and then compute the confusion matrix.  Next, we re-sort the columns of the matrix in such a way so that the sum of diagonal elements in the new confusion matrix is maximized. For the co-segmentation of the comparison methods, we set the clusters number is the same as the number of ground-truth classes (seven class). For IDUS, we set the dimension of the network softmax output as seven. Therefore, we use a confusion matrix with size $7\times7$ as the criterion for evaluating different methods.

\figurename~\ref{fig:confusion_matrix} shows the confusion matrices of IDUS and the comparison methods. As shown in the figure, IDUS results in superior co-segmentation results (\figurename~\ref{fig:confusion_matrix}c) over the comparison methods (\figurename~\ref{fig:confusion_matrix}a and b). Upon examining the confusion matrices, we notice some comparison methods do not provide enough discrimination ability to differentiate certain classes well. For example, Shadow (SH) and Dark Sand (DS) are clustered into the same class in Lianantonakis, \textit{et al.} \cite{lianantonakis2007sidescan} (\figurename~\ref{fig:confusion_matrix}a). However, such misclassification problems are diminished by IDUS (\figurename~\ref{fig:confusion_matrix}c).
% whole section should be a page

% The dataset we use to test the performance of IDUS is composed of eighty images collected from a high-frequency SAS system which are semi-labeled (i.e. not every pixel is labeled in every image). The images contain seven distinct labeled seabed areas: shadow (SH), dark sand (DS), bright sand (BS), seagrass (SG), large sand ripple (SR), rock (RK), and small sand ripple (SR). We separate the dataset into a train-validation split by ratio $7:3$. We use U-Net \cite{unet} as a backbone network in IDUS. We compare our method with two state-of-the-art unsupervised segmentation methods Wavelet \cite{williams2009unsupervised} and Sobel+HoG+LBP\cite{Zare2017PossibilisticFL}. Figure \ref{fig:exp1} shows the confusion matrices of seven classes for IDUS against the two comparison methods, Sobel-HoG-LBP, and Wavelets. The mean accuracy of IDUS ($0.7$) on all classes is largely beyond the Wavelet ($0.57$) and Sobel+Hog+LBP ($0.52$) methods demonstrating the improved results of our method.

% In our full paper, we will demonstrate how this powerful method can be combined with a supervised method to improve the network performance in small sample cases. An investigation of our own method with different clustering groups will also be performed to see how this method differentiates the textures that humans are hard to classify.

\section{Conclusion}
% one or two paragraphs
% a sentence about future work
% exteding to supervised...
We propose an unsupervised method for SAS seabed image segmentation which works by iteratively clustering on superpixels while training a deep model. Our proposed method (iterative, deep, and unsupervised segmentation (IDUS)) does not require large amounts of labeled training data which are difficult to obtain for SAS. We show that IDUS obtains state-of-the-art performance on a real-world SAS dataset and show performance benefits by comparing against existing methods. 

In future work, we will extend IDUS to combine with supervised or semi-supervised training data.

\section{Acknowledgements}
This work was supported by the Office of Naval Research via grant N00014-19-1-2513. The authors thank the Naval Surface Warfare Center - Panama City Division for providing the data used in this work.
% one sentence

% \begin{table}[]
% \begin{tabular}{|l|l|l|l|l|l|l|l|}
% \hline
% Method & SH    & DS    & BS    & SG    & SR    & RK  & LR    \\ \hline
% U-Net  & 0.898 & 0.968 & 0.975 & 0.974 & 0.985 & 1.0 & 0.999 \\ \hline
% IDUS   & 0.282 & 0.881 & 0.655 & 0.261 & 0.858 & 1.0 & 0.96  \\ \hline
% \end{tabular}
%     \caption{The pixel accuracy of each class for IDUS and U-Net in the second experiment.}
%     \label{tab:exp2} 
% \end{table}

% ==================================================================================================
% Conclusion
% ==================================================================================================

\bibliographystyle{IEEEtran}
\bibliography{paper}

\end{document}